\documentclass[journal,twoside,web]{ieeecolor}
\usepackage{generic}
\usepackage{cite}
\usepackage{amsmath,amssymb,amsfonts}
\usepackage{algorithmic}
\usepackage{graphicx}
\usepackage{textcomp}
\usepackage{booktabs}
\usepackage{diagbox}
\usepackage{cuted}
\def\BibTeX{{\rm B\kern-.05em{\sc i\kern-.025em b}\kern-.08em
    T\kern-.1667em\lower.7ex\hbox{E}\kern-.125emX}}
\begin{document}
\title{LLMs-guided adaptive compensator: Bringing Adaptivity to Automatic Control Systems with Large Language Models}
\author{Zhongchao Zhou, Yuxi Lu  \IEEEmembership{Member, IEEE}, Yaonan Zhu, Yifei Zhao, Bin He \IEEEmembership{Senior Member, IEEE},  Liang He, Wenwei Yu and Yusuke Iwasawa 
\thanks{This work was supported in xxx. }
\thanks{Zhongchao Zhou, Yaonan Zhu and Yusuke Iwasawa are with the School of
Engineering, The University of Tokyo, Tokyo, 113-0033, Japan. (e-mail: zhouzhongchao@outlook.com, yaonan.zhu@weblab.t.u-tokyo.ac.jp, iwasawa@weblab.t.u-tokyo.ac.jp).}
\thanks{Yuxi Lu, Yifei Zhao and Bin He are with the Shanghai Research Institute for Intelligent Autonomous Systems, Tongji University, Shanghai 201210, China (e-mail: yuxilu@tongji.edu.cn, yifeizhao@tongji.edu.cn, hebin@tongji.edu.cn).}
\thanks{Liang He is with the Institute of Biomedical Engineering (IBME), University of Oxford,
Oxford, UK, OX3 7DQ  (e-mail: liang.he@eng.ox.ac.uk).}
\thanks{Wenwei yu is with Graduate School of Science and Engineering and Center for Frontier Medical Engineering, Chiba University, Chiba, 2638522, Japan  (e-mail: yuwill@faculty.chiba-u.jp).}
}

\maketitle

\begin{abstract}
With rapid advances in code generation, reasoning, and problem-solving, Large Language Models (LLMs) are increasingly applied in robotics, most existing work focuses on high-level tasks such as task decomposition. A few studies have explored the use of LLMs in feedback controller design, however, these efforts are restricted to overly simplified systems, fixed-structure gain tuning, and lack real-world validation. To further investigate LLMs in automatic control, this work targets a key subfield: adaptive control. Inspired by the framework of model reference adaptive control (MRAC), we propose an LLMs-guided adaptive compensator framework that avoids designing controllers from scratch. Instead, the LLMs are prompted using the discrepancies between an unknown system and a reference system to design a compensator that aligns the response of the unknown system with that of the reference, thereby achieving adaptivity. Experiments evaluate five methods—LLM-guided adaptive compensator, LLM-guided adaptive controller, indirect adaptive control, learning-based adaptive control, and MRAC—on soft and humanoid robots, in both simulated and real-world environments. Results show that the LLMs-guided adaptive compensator outperforms traditional adaptive controllers and significantly reduces reasoning complexity compared to the LLMs-guided adaptive controller. The Lyapunov-based analysis and reasoning-path inspection demonstrate that the LLMs-guided adaptive compensator enables a more structured design process by transforming mathematical derivation into a reasoning task, while exhibiting strong generalizability, adaptability, and robustness. This study opens a new direction for applying LLMs in the field of automatic control, offering greater deployability and practicality compared to vision-language models.
\end{abstract}


\section{Introduction}
\label{sec:introduction}
\IEEEPARstart{L}{arge} Language Models (LLMs) have demonstrated remarkable capabilities across a wide range of domains, including code generation \cite{jiang2024survey}, advanced mathematics\cite{ahn2024large}, and logical reasoning through chain-of-thought (CoT) prompting \cite{wei2022chain}. In robotics, recent research has begun exploring how LLMs can be used high-level perception and language understanding, enabling robots to execute long-horizon tasks based on natural language instructions. As shown in Figure 1-A, a typical approach involves using the LLMs to parse instructions such as “put the wine glass in the kitchen cabinet” and generate corresponding sub-task sequences for manipulation \cite{zeng2023large}. Moreover, for more complex tasks which involve temporal dependencies and multi-step reasoning, LLMs-based approaches combined with human-robot collaboration have also been shown effective\cite{liu2024enhancing}.  In addition to task planning, recent efforts have also extended LLMs toward vision-language-action integration, allowing robots to perceive, understand, and act in the physical world using multimodal inputs\cite{sapkota2025vision}. Concurrently, researchers are also building robot-specific foundation models and multimodal datasets, such as VIMA, Gemini Robotics Model and Pi0, to improve generalization and real-world applicability in embodied AI \cite{jiang2022vima}, \cite{black2024pi_0}, \cite{team2025gemini}. \\
As shown in Figure1-B, in robotic control, a key branch alongside motion planning is automatic control. This area focuses on the continuous-time dynamics of robotic systems, aiming to design feedback controllers based on physical (particularly dynamic) models to regulate and stabilize states such as position, velocity, and force~\cite{albertos2010feedback}. Unlike planning, where trajectory generation allows for offline long computation time, automatic control prioritizes real-time feedback and dynamic regulation. It typically relies on equations of motion (based on Lagrangian or Newton-Euler formulations) to inform closed-loop controller design.

\begin{figure}
    \centering
    \includegraphics[width=1\linewidth]{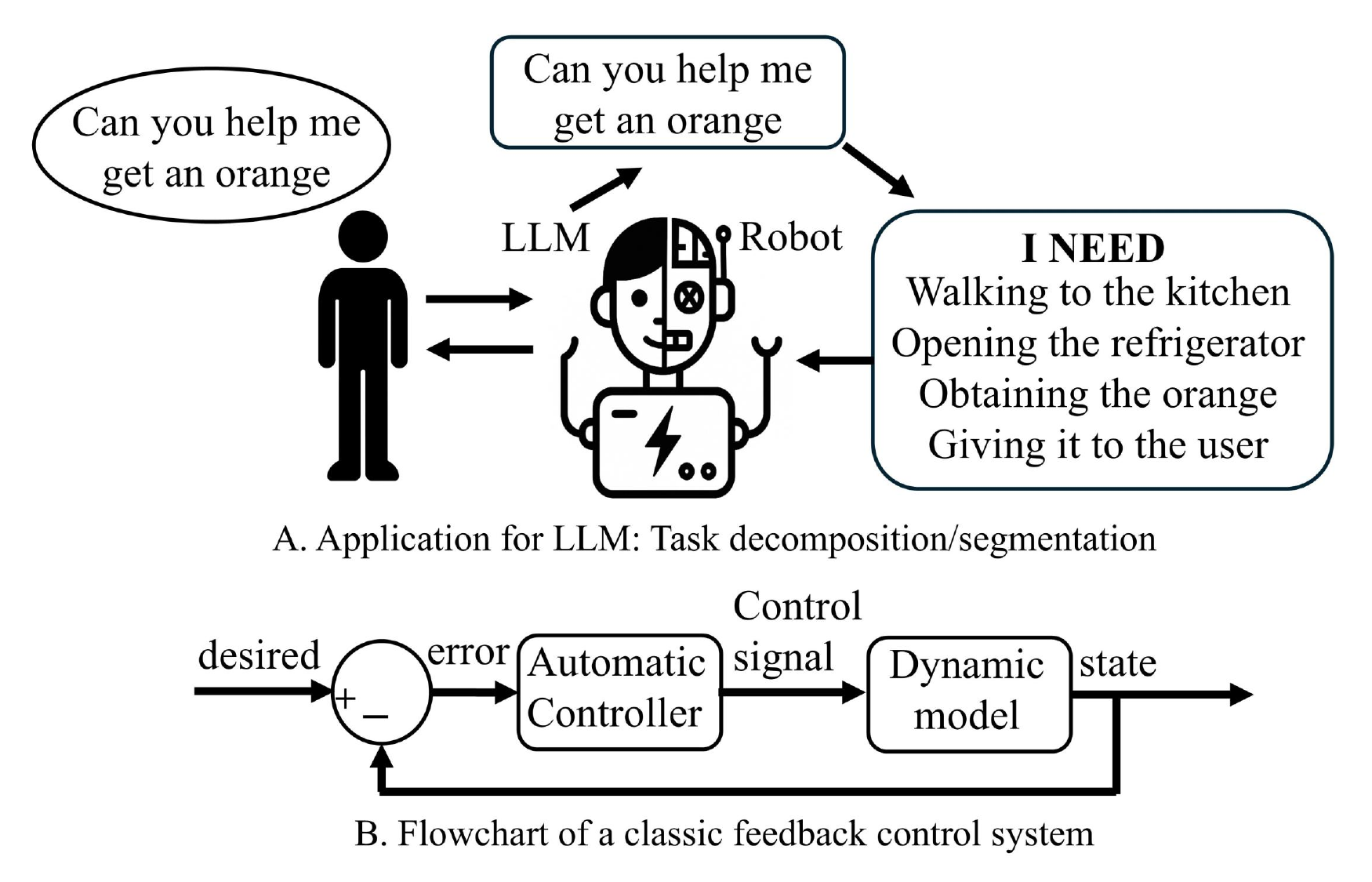}
    \caption{Trends in LLM-based robotics and feedback control}
    \label{fig:enter-label}
\end{figure}

\subsection{Related research – LLM for Automatic control}
In the domain of automatic control, the application of LLMs to controller design remains extremely limited. According to our investigation, only a few studies have started to explore this direction \cite{kevian2024capabilities}, \cite{guo2024controlagent}, \cite{narimani2025agenticcontrol}, \cite{tohma2025smartcontrol}.
These works attempt to showcase the potential of LLMs in control design by leveraging their mathematical reasoning abilities. Specifically, the LLMs are provided with the dynamic model of the robot/system (the transfer function of the system) and a set of control performance requirements (e.g. stability, phase margin, and settling time). LLMs are guided through iterative prompts to design feedback controllers that meet these specifications. Experimental results suggest that LLMs outperform traditional auto-tuning tools PIDtune (provided by Matlab) in terms of both average successful rate and aggregate success rate. Some studies have also proposed software frameworks that integrate LLMs to tune PID controller gains. The underlying principle is also to provide the LLMs with control performance requirements, based on which it adjusts the PID gains to achieve the target performance. 

\subsection{Related research – Adaptive control and LLM for Adaptive control}

Adaptive control has long been an important direction in the field of automatic control systems. Its core objective is to maintain desirable control performance and system stability in the presence of uncertainties (e.g. unknown or time-varying parameters in the robot dynamic system) \cite{sastry2011adaptive}. Traditional adaptive control methods are generally categorized into two types: indirect adaptive control and direct adaptive control\cite{hosseini2018online}. In \cite{zahedifar2025llm} authors first time proposes a control architecture that integrates LLMs with traditional  adaptive controllers in an attempt to address performance degradation caused by trajectory variations or external disturbances.\\

Despite the growing interest in applying LLMs to automatic control, current research still faces several critical limitations: \\

1. \textbf{Over-simplified system:} Most existing studies are confined to linear time-invariant systems with known transfer functions. Even in \cite{zahedifar2025llm}, the experiments are conducted on commercial robotic arms with well-established dynamic models. In that study, the “unmodeled dynamics”  is essentially treated as an external disturbances, which is theoretically closer to robustness enhancement than true adaptive control. Therefore, no prior work has explored whether LLMs can still achieve automatic control in the presence of highly complex, nonlinear, time-varying, or inherently unmodelable systems. Even the fundamental question of how to construct meaningful prompts for such systems remains entirely unaddressed.\\

2. \textbf{Limited to gain tuning in fixed control structures:} Nearly all existing approaches use LLMs merely to tune control gains of pre-defined controllers, without the ability to generate novel control laws or explore structural-level design. Even in \cite{zahedifar2025llm}, the role of the LLMs ultimately reduce to adjusting gains in a known adaptive controller. If confined to this level of use, LLMs offer little advantage over conventional optimization techniques or manual tuning, and their potential for reasoning and generation remains underutilized. Crucially, there has been no work comparing LLMs-generated controllers with those designed by classical control theory in terms of design complexity, structural formulation, and final performance.\\

3. \textbf{Lack of real-world validation and analyze:} To date, no studies have deployed fully LLMs-generated controllers on real-world robotic platforms. Most evaluations remain confined to simulation environments, likely due to concerns about the generalizability, and safety of the generated controller. Without analysis of the LLM’s reasoning path or validation of its robustness under real-world uncertainties, the practical feasibility and engineering value of such controllers remain uncertain.\\

To address the challenges of applying LLMs in automatic control, especially in adaptive controllers, we propose a new perspective for utilizing LLMs in the field of adaptive control. Rather than expecting LLMs to design a complete feedback adaptive controller from scratch or merely tune an existing feedback controller. Instead, we proposed an LLMs-guided adaptive compensator that enhances the adaptivity of an existing feedback controller.  This framework is inspired by the concept of Model Reference Adaptive Control (MRAC). The process of the LLMs-guided adaptive compensator can be summarized as follows: the user first defines a reference system with known dynamics and, through trial and error, obtains a reference response. By prompting the LLM with both the reference response and the observed response of the unknown system, the model is guided to design a compensator. The objective of this compensator is to adjust the unknown system's response to closely match that of the reference system, thereby enhancing the adaptivity of the existing feedback controller. The proposed LLMs-guided adaptive compensator was systematically compared against several traditional adaptive controllers, including indirect model-based adaptive control, MRAC, learning-based adaptive control, and an LLMs-guided adaptive controller in which the LLM is prompted to directly design adaptive controller. Moreover, the experiments were tested on two representative platforms: a McKibben Pneumatic Artificial Muscles (PAMs)-driven robotic arm and a motor-driven humanoid robot. Both simulation and prototype experiments consistently demonstrate that the LLMs-guided adaptive compensator achieves superior response performance and ease of use compared to traditional adaptive controllers. Beyond validating its feasibility through experiments, we further conducted Lyapunov-based stability analysis and reasoning-path inspections. The analyses demonstrate that the proposed compensator can be generally applied to a broad class of unknown systems with similar structural characteristics, while requiring significantly lower reasoning complexity. It also yields clearer and more interpretable CoT. Notably, the LLM-guided adaptive compensator effectively leverages the reasoning capabilities of LLMs—rather than their symbolic derivation skills—highlighting its practicality and potential for scalable deployment in real-world control tasks.\\
Our contribution could be summarized as follows:\\
1. We propose, for the first time, a novel framework that leverages LLMs to enhance the adaptivity of existing feedback controllers through compensator design for complex systems. Moreover, this is also the first demonstration that, compared with several representative traditional adaptive controllers, the LLMs-guided compensator consistently achieves superior performance\\
2. We first time theoretically analyze the proposed LLMs-guided adaptive compensator in terms of generalizability, adaptability, and robustness via Lyapunov-based stability analysis and reasoning-path inspection. We also provide analysis to explain its advantages over directly designing controllers via LLMs.\\
3. We conduct the first time experiments on real-world prototype environments to validating the practical applicability .

\begin{figure*}
    \centering
    \includegraphics[width=1\linewidth]{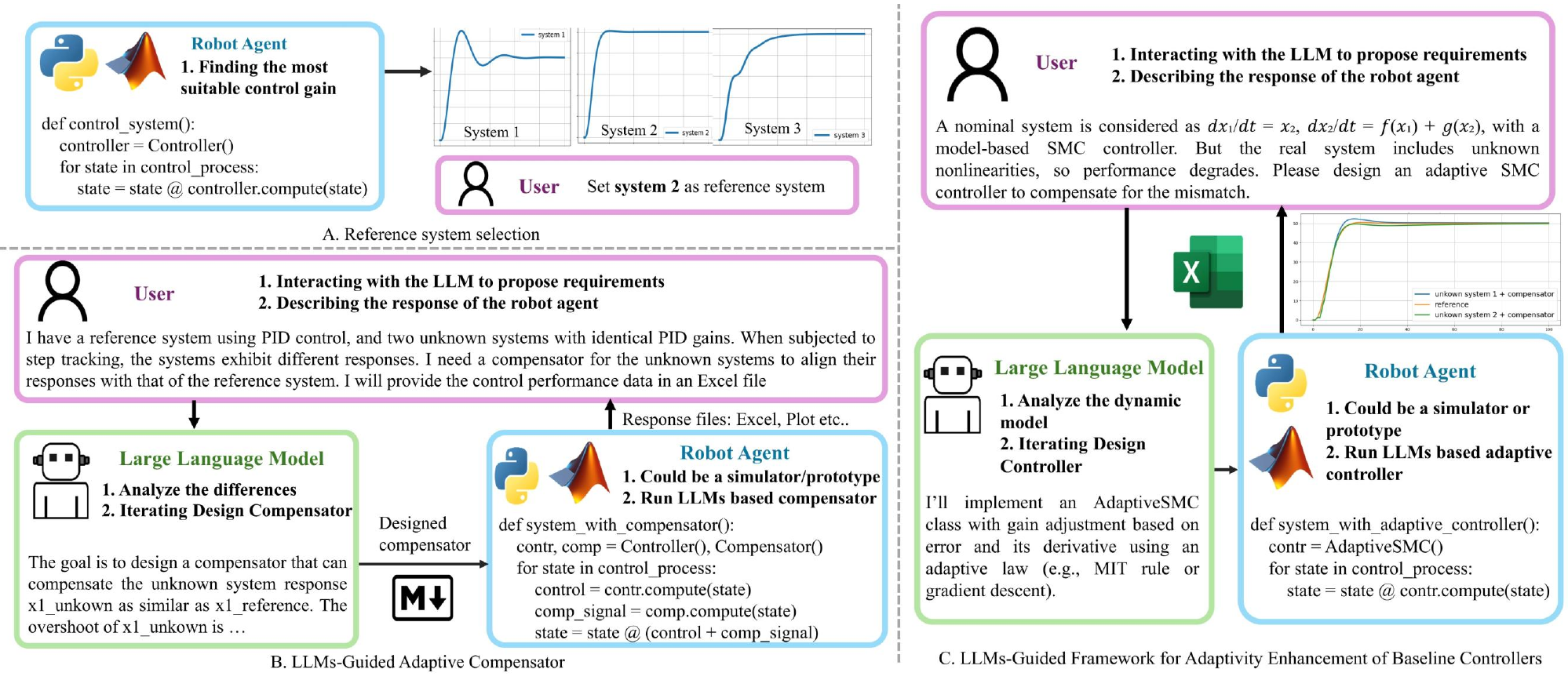}
    \caption{Flowchart of LLMs-Guided adaptive compensator and LLMs-Guided adaptive controller }
    \label{fig:enter-label}
\end{figure*}

\section{Methods}

\subsection{LLMs-guided adaptive compensator}
\textbf{Reference system selection:}\\
As illustrated in Figure2A, the first step in the compensator design process is to define an appropriate reference system that serves as the desired target for control behavior. While the problem of selecting reference models has been extensively explored in prior work\cite{arabi2020safety, zhou2024gan, stepanyan2010mrac}, this study does not aim to evaluate alternative reference model designs in depth. Instead, we adopt a simple but widely accepted strategy: the reference system is constructed using a well-modeled system with fixed controller gains, and its response exhibits desirable characteristics such as smooth response, fast convergence, and no overshoot. This configuration is manually selected through a trial-and-error process~\cite{shen2002adaptive}. Specifically, System~2 in Figure~2A exhibits superior transient dynamics. The LLMs-guided adaptive compensator aims to align the responses of Systems~1 and~3 with that of System~2. \\

\textbf{Compensator Design:}\\
The compensator design process, shown in Figure~3B, involves natural language interaction between the user and the LLM. The user specifies the task objective and supplies the responses of both the reference and unknown systems. Qualitative descriptors (e.g., “sluggish response”, “high overshoot”) may also be included to enrich the prompt.

Upon receiving the task prompt $P$, the LLMs function as a nonlinear mapping:
\[
\mathcal{L} : \mathcal{P} \rightarrow \mathbb{R}^m
\]
where $\mathcal{P}$ is the input prompt space containing a mixture of natural language instructions, reference and unknown system responses, and possibly user preferences. This mapping $\mathcal{L}$ is inherently non-parametric, implicit, and multi-modal, relying on high-dimensional embeddings learned from both linguistic and dynamical priors.

The output $\phi_c$ corresponds to a synthesized compensator, which is inferred by the LLMs as a compensation strategy and can be explicitly expressed in functional or code form. It is designed to augment an existing feedback controller $u_0(t)$. The total control input applied to the unknown system is formulated as:

\[
u(t) = u_0(t) + \phi_c(u_\mathrm{base}, y_\mathrm{desired}, y(t), y_r(t), k)
\]

Here, $u_0(t)$ denotes the existing feedback controller, which remains fixed during the compensator design process. Let $y(t)$ represent the state of the unknown system, $y_r(t)$ the state of the reference system, $k$ denotes the discrete time index at continuous time $t$, $u_{\text{base}}$ the output of the existing controller, and $y_{\text{desired}}$ the desired system output. The compensator $\phi_c(\cdot)$ is designed to correct deficiencies in $u_0(t)$, guiding the system response $y(t)$ to more closely match the reference trajectory $y_r(t)$. 
After that the LLMs-guided adaptive compensator is passed to a Robot Agent, which can be either a simulation platform or a physical robotic system. After each round of control execution (indexed by $i$), the updated system response $y^{(i)}(t)$ from the unknown system is returned to the user. Based on the observed performance, the user provides feedback—typically also denoted as $y^{(i)}(t)$—which is then interpreted by the LLM to generate a refined compensation strategy.This process forms a closed-loop interactive cycle, where the compensator is iteratively improved across rounds. The update rule can be formalized as:
\[
\phi_c^{(i+1)} = \phi_c^{(i)} + \Delta^{(i)}, \quad \Delta^{(i)} := \mathcal{L}(y^{(i)}(t))
\]

\subsection{LLMs-guided for adaptive controller }
This framework is illustrated in Figure~2C. It represents the most straightforward and intuitive approach for leveraging LLMs in control design, and has also been adopted in related studies such as~\cite{guo2024controlagent},~\cite{narimani2025agenticcontrol}. While the overall structure of this approach resembles that of the compensator design process, there are key differences in the type of input provided to the LLMs and the way control logic is constructed.

In this framework, the user does not provide system responses directly. Instead, the LLMs receive an explicit mathematical description of the system dynamics, typically in the form of differential equations, transfer functions, or state-space models such as:
\[
\dot{x}(t) = f(x(t), u(t)), \quad y(t) = h(x(t)),
\]

The form of an existing controller may also be included in the prompt. Based on this information, the LLM is tasked with designing an adaptive controller from scratch, resulting in a new control law:
\[
u(t) = u_{\mathrm{adapt}}(t),
\]
where \( u_{\mathrm{adapt}}(t) \) is synthesized entirely by the LLMs and may incorporate adaptive elements such as online parameter estimation, gain scheduling, or Lyapunov-based stability mechanisms.

Once the controller is generated, it is deployed within the Robot Agent and then follows an iterative refinement loop, where the controller is progressively improved based on observed system response and user feedback.

\section{LYAPUNOV-BASED GENERALIZATION ANALYIS OF LLMS-GUIDED ADAPTIVE COMPENSATOR}

It is hypothesized that LLMs-guided adaptive compensator may overfit to the specific system described in the prompt, leading to compensators that perform well only on the system described in the prompt, rather than generalizing across diverse dynamics. To examine whether such compensators exhibit theoretical generalizability, a Lyapunov-based analysis is employed. In particular, if a common Lyapunov function can be established that guarantees stability across a family of systems—even when the state or input matrices differ—the compensator could be regarded as generalizable for that class \cite{loria2018strict}. \\
To verify that, a nonlinear time-invariant system is first established as the reference system. Its state-space representation is provided in Equation (1).

\begin{equation}
\begin{bmatrix} \dot{x}_1 \\ \dot{x}_2 \end{bmatrix}
=
\begin{bmatrix} 0 & 1 \\ -5 & -3 \end{bmatrix}
\begin{bmatrix} x_1 \\ x_2 \end{bmatrix}
+
\begin{bmatrix} 0 \\ 1 \end{bmatrix} u
\end{equation}

where $x_1$ denotes the system position, $x_2$ represents the velocity, and $u$ is the control input. For the reference system, a model-based sliding mode controller (SMC) is designed and shown as follows:

\begin{equation}
u = \frac{1}{g} \left( -f - \lambda \dot{e} - k\,\text{sign}(s) - \gamma \int s\, dt \right)
\end{equation}

where $f(\cdot)$ is the state matrix, $g(\cdot)$ is the input matrix, and $\lambda$, $k$, and $\gamma$ are control gains, all strictly positive.  $e = x_1 - x_{\text{1d}}$ represents the tracking error and $x_{\text{1d}}$ denotes the desired position. The function $\text{sign}(\cdot)$ denotes the sign function, and $s$ is the sliding surface. The detailed Lyapunov stability proof is provided in Supplementary Material~1, which demonstrates that the proposed SMC controller guarantees asymptotic stability.

\begin{figure*}
    \centering
    \includegraphics[width=1\linewidth]{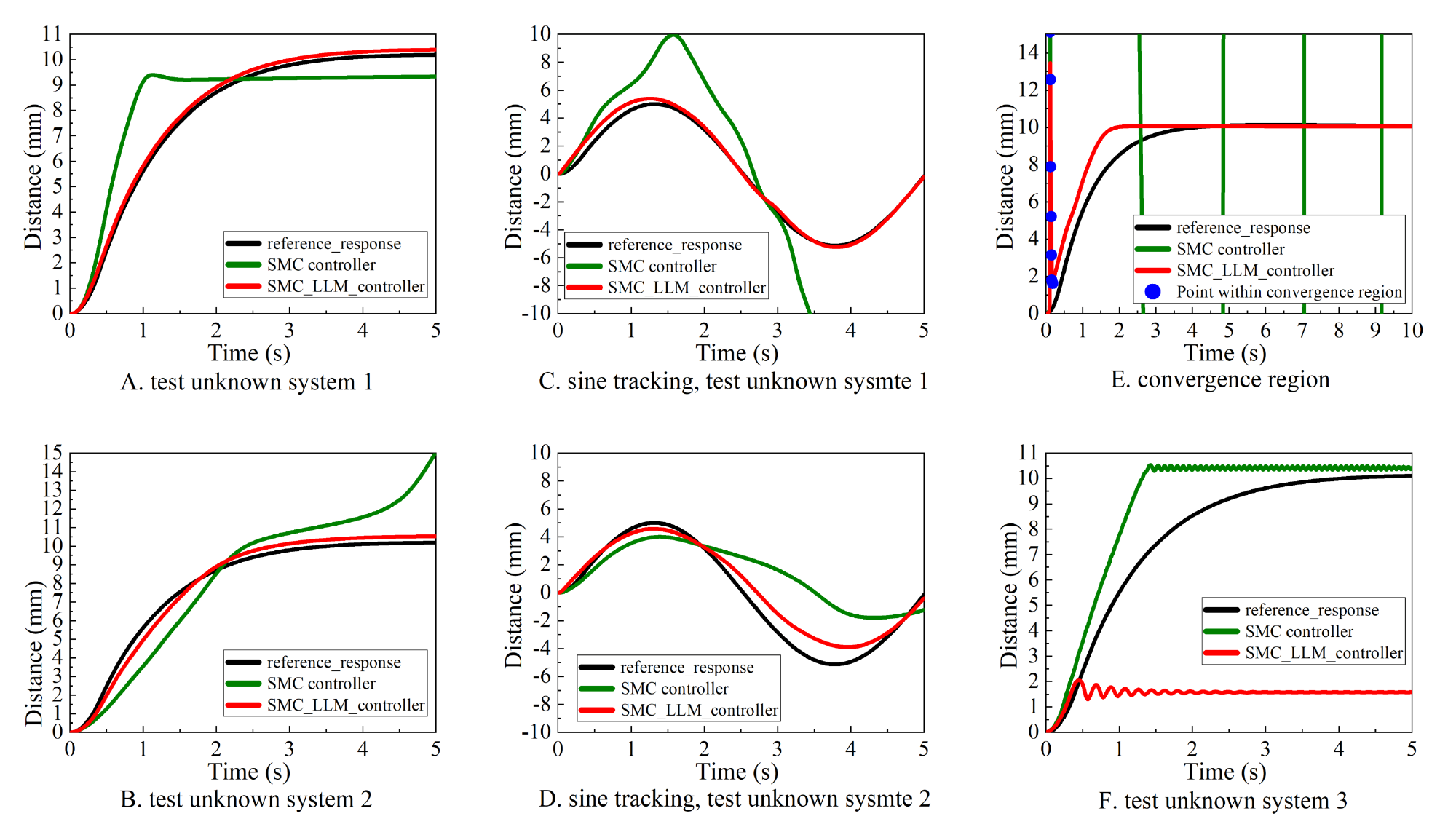}
    \caption{LLMs-guided adaptive compensator validation }
    \label{fig:enter-label}
\end{figure*}

The designed SMC controller was applied to two different systems (defined as Unknown System 1 and 2), while the functions $f(\cdot)$ and $g(\cdot)$ inside the controller were kept identical to the parameterization given in Equation~(1).

Unknown System 1 is described by the following equations:
\begin{equation}
\begin{aligned}
\begin{bmatrix}
\dot{x}_1 \\
\dot{x}_2
\end{bmatrix}
&=
\begin{bmatrix}
0 & 1 \\
-3.5 + 3 \sin(0.5x_1) & -3 + 3 \cos(0.2x_2)
\end{bmatrix}
\begin{bmatrix}
x_1 \\
x_2
\end{bmatrix} \\
&\quad +
\begin{bmatrix}
0 \\
2 + 2\sin(x_1)
\end{bmatrix} u
+
\begin{bmatrix}
0 \\
-0.2 \sin(x_1) + 0.05 \xi(t)
\end{bmatrix}
\label{eq:sys10}
\end{aligned}
\end{equation}

Unknown System 2 is described by the following equations:
\begin{equation}
\begin{aligned}
\begin{bmatrix}
\dot{x}_1 \\
\dot{x}_2
\end{bmatrix}
&=
\begin{bmatrix}
0 & 1 \\
-4 + 2 \sin(x_1) & -10 + 3 \tanh(0.2x_2)
\end{bmatrix}
\begin{bmatrix}
x_1 \\
x_2
\end{bmatrix} \\
&\quad +
\begin{bmatrix}
0 \\
0.8 + 0.2 \cos(x_1)
\end{bmatrix} u
+
\begin{bmatrix}
0 \\
-0.2 \sin(x_1) + 0.5 \xi(t)
\end{bmatrix}
\label{eq:sys11}
\end{aligned}
\end{equation}

An LLMs-guided adaptive compensator is prompted in a \textbf{zero-shot} manner to enable both Unknown System 1 and Unknown System 2 to track the reference system. The closed-form expression is given in Equation~\eqref{eq:ucomp}, with the full reasoning and step-by-step design process provided in the Supplementary Material 1.

\begin{equation}
u_{\text{comp}} = k_p e_1 + (k_d + k_v) e_2 + k_i I
\label{eq:ucomp}
\end{equation}

Here, $e_1$ denotes the position error between the reference system and the unknown system, $e_2$ represents the corresponding velocity error, and $I$ is defined as the time integral of the tracking error for the unknown system.

To evaluate the generability of the LLMs-guided adaptive compensator, a Lyapunov-based analysis is conducted. Equation~(1), Equation~\eqref{eq:sys10}, Equation~(4) are reformulated in the following form:

\begin{equation}
\begin{bmatrix}
\dot{x}_{1,r} \\
\dot{x}_{2,r}
\end{bmatrix}
=
\begin{bmatrix}
x_{2,r} \\
-5x_{1,r} - 3x_{2,r} + u_r
\end{bmatrix}
\label{eq:refsystem}
\end{equation}

\begin{equation}
\begin{bmatrix}
\dot{x}_{1,u} \\
\dot{x}_{2,u}
\end{bmatrix}
=
\begin{bmatrix}
x_{2,u} \\
a_{21} x_{1,u} + a_{22} x_{2,u} + b u_u + d(t)
\end{bmatrix}
\label{eq:unknowndyn}
\end{equation}

where
\[
-6.5 \leq a_{21} = -3.5 + 3 \sin(x_{1,u}) \leq -0.5
\]
\[
-6 \leq a_{22} = -3 + 3 \cos(0.2x_{2,u}) \leq 0
\]
\[
0 \leq b = 2 + 2\sin(x_{1,u}) \leq 4
\]
\[
|d(t)| = |-0.2 \sin(x_1) + 0.05\xi(t)| \leq 0.25
\]

The control input is:
\[
u_u = u_{\text{base}} + u_{\text{comp}}
\]

The error dynamics are defined as:
\begin{align}
\dot{e}_1 &= \dot{x}_{1,r} - \dot{x}_{1,u} = e_2 \label{eq:e1dot} \\
\dot{e}_2 &= \dot{x}_{2,r} - \dot{x}_{2,u} \label{eq:e2dot}
\end{align}

By substituting Equations~\eqref{eq:refsystem} and \eqref{eq:unknowndyn} into \eqref{eq:e2dot}, we get:

\begin{equation}
\dot{e}_2 = -5x_{1,r} - 3x_{2,r} + u_r - a_{21}x_{1,u} - a_{22}x_{2,u} - b u_u - d(t)
\label{eq:e2dotfinal}
\end{equation}

The Lyapunov candidate function is defined as:
\begin{equation}
V = \frac{1}{2}(e_1^2 + e_2^2)
\label{eq:lyap2}
\end{equation}

Taking its derivative and substituting \eqref{eq:e2dotfinal} yields:
\begin{equation}
\dot{V} = e_1 \dot{e}_1 + e_2 \dot{e}_2 = e_1 e_2 - b e_2 u_{\text{comp}} + e_2 \Delta_{\text{fixed}}
\label{eq:vdot}
\end{equation}

where $\Delta_{\text{fixed}}$ denotes the system differences:
\begin{equation}
\Delta_{\text{fixed}} = a_{21}x_{1,u} + a_{22}x_{2,u} - 5x_{1,r} - 3x_{2,r} + b u_{\text{base}} - u_r + d(t)
\label{eq:delta}
\end{equation}

By the triangle inequality, the upper bound for $\Delta_{\text{fixed}}$ is:

\begin{equation}
\begin{aligned}
|\Delta_{\text{fixed}}| \leq\ & |a_{21}||x_{1,u}| + |a_{22}||x_{2,u}| + 5|x_{1,r}| \\
& + 3|x_{2,r}| + b|u_{\text{base}}| + |u_r| + |d(t)|
\end{aligned}
\label{eq:delta_bound}
\end{equation}

The system parameters and controller gains resulting from the LLMs-guided adaptive compensator design are as follows:
\[
k_p = 20.0, \quad k_d = 10.0, \quad k_v = 8.5, \quad k_i = 1, \quad b = 1
\]

Substituting these into \eqref{eq:vdot}:
\begin{equation}
\dot{V} = e_2(-19e_1 - 18.5e_2 - I + |\Delta_{\text{fixed}}|)
\label{eq:vdotfinal}
\end{equation}

To ensure $\dot{V} < 0$, three cases are analyzed:

**Case 1**: $e_2 > 0$:
\begin{equation}
-19e_1 - 18.5e_2 - I + 274.45 < 0
\label{eq:case1}
\end{equation}

**Case 2**: $e_2 < 0$:
\begin{equation}
-19e_1 - 18.5e_2 - I - 274.45 > 0
\label{eq:case2}
\end{equation}

**Case 3**: $e_2 = 0$:
\[
\dot{V} = 0
\]

If the integral term is neglected (i.e., $I = 0$) and the expression is simplified, then when $e_2 > 0$::
\begin{equation}
1.02 e_1 + e_2 > 15
\label{eq:simple}
\end{equation}

When $e_2 < 0$, the following condition must be satisfied:
\begin{equation}
1.02e_1 + e_2 < -15
\label{eq:condition_negative}
\end{equation}

Therefore, the LLMs-guided adaptive compensator combined with the existing controller exhibits \textbf{regionally asymptotic stability}. That is, the compensator exhibits effectiveness within a bounded region defined by specific error conditions, thereby gradually compensating the response of the unknown system closer to that of the reference system.

As long as the error dynamics structure of the unknown system closely aligns with Equation~(1), namely:
\begin{equation}
\dot{x} = A x + B u
\label{eq:stateform}
\end{equation}
regardless of whether $A$ and $B$ are linear or time-invariant, the compensator remains effective in reducing the tracking error.The key variable is the radius of the convergence region, which depends on the dynamic discrepancy between the unknown and reference systems. A larger discrepancy requires a greater initial error for the compensator to become active, shifting the convergence region farther from the origin; a smaller discrepancy enables compensation within a narrower error range.\\
Although the Lyapunov-based analysis is conducted on a single system, its insights are argued to be generalizable due to the zero-shot nature of the design process. The compensator was generated solely through zero-shot prompting—without relying on predefined control templates, handcrafted structures, or manual tuning. Consequently, regardless of the user or specific system, similar reasoning logic is applied as long as the input follows the same formulation. To validate the effectiveness of the LLMs-guided adaptive compensator, systematic experiments were conducted on multiple systems, as shown in Figure 3. Figures 3 A and B evaluate two novel nonlinear time-varying systems, distinct from those in Equations~(4) and (5). In all systems, the compensator successfully achieved the desired control objectives, whether the unknown system responded faster or slower than the reference. Figures 3 C and D present sinusoidal tracking tasks, where the compensator maintained excellent performance. In Figure 3E, a disturbance was applied at $t = 0.02$ s, shifting the state to 13~mm. Without compensation, the system diverged; with the compensator, the state returned to stability and resumed tracking. The blue point marks the disturbed state, lying within the region theoretically proven (via Equation~(16)) to ensure convergence under Lyapunov stability. The observed recovery aligns with theory, validating robustness within the defined convergence region. Finally, Figure 3F tests an unknown system (Unknown System 3) that deviates structurally from Equation~\eqref{eq:stateform}. Although its complexity is comparable to earlier systems, the violation of structural assumptions renders the compensator ineffective, resulting in poor tracking performance.

\begin{figure}
    \centering
    \includegraphics[width=1\linewidth]{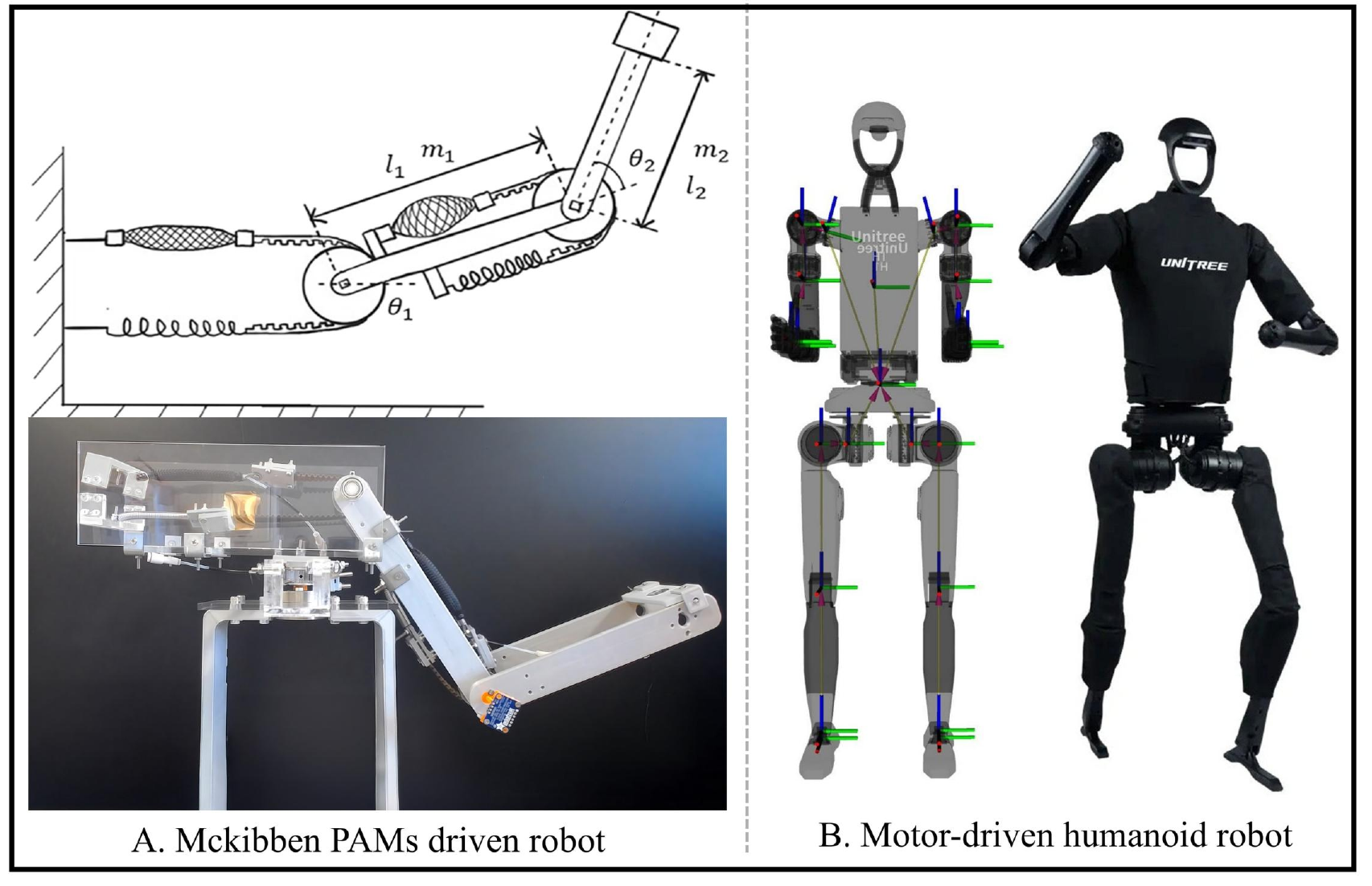}
    \caption{Flowchart of LLMs-Guided adaptive compensator and LLMs-Guided adaptive controller }
    \label{fig:enter-label}
\end{figure}

\section{EXPERIMENT SETTING}
\subsection{Robot Control test}
\textbf{Type 1 (T-1):} Mckibben PAMs-Driven robot arm (1-DoF, 2-DoF Simulation \& Prototype)\\

To evaluate the effectiveness of the proposed LLMs-guided adaptive compensator on complex systems and to compare it with several traditional adaptive controllers, a two-degree-of-freedom (DoF) Mckibben PAMs-driven robot arm was selected as the test platform, as illustrated in Figure 4A. PAMs are typical form of soft actuators, inherently exhibit strong nonlinearity, hysteresis, and air compressibility \cite{liu2024fixed}. When extended to a 2-DoF configuration, additional dynamic coupling between joints further increases control complexity. In all experiments, the external load was treated as an unknown parameter. The dynamic model of the system was derived and validated in \cite{zhou2022role}. All other experimental conditions were kept identical across the different controller to ensure fair comparison.
 
1. Direct model based adaptive controller via Lyapunov stability theory:  A classical adaptive controller was established using Lyapunov stability theory. The Lyapunov function incorporated both state errors and external load estimation error, allowing for simultaneous trajectory tracking and load adaptation. As the control law was derived from system dynamics and updated via online estimation, this approach represents a standard model-based adaptive strategy. The full stability proof is provided in Supplementary Material 2.\\
2. Model reference adaptive controller: An MRAC was implemented using Lyapunov theory, with a pre-defined reference system and adaptive laws based on state and gain estimation errors. The goal was to drive the unknown-load system to asymptotically track the reference response. Stability analysis follows prior work in \cite{zhou2024simulation}, and the reference controller (SMC-based) is detailed in \cite{zhou2022optimization}.\\
3. learning based adaptive controller: A GAN-based PID controller(G-PID) was implemented as a data-driven adaptive controller. The generator served as a compensator to modify the control input, while the discriminator evaluated the discrepancy between the unknown and a reference system. Through adversarial training, the controller gradually learned to align the unknown system response with the reference, without relying on explicit physical modeling \cite{zhou2024gan}. \\

\textbf{Type 2 (T-2):} Humanoid Platforms (Prototype)\\
To evaluate the applicability of the LLMs-guided adaptive compensator beyond conventional robotic systems, we further deployed it on a commercial humanoid robot platform (Unitree H1\_2), as illustrated in Figure~4B. Due to inevitable modeling discrepancies between the simulation and real-world platforms, using identical Proportional-Derivative (PD) controller in both environments led to noticeably different response. This makes the humanoid scenario a representative case of a sim-to-real control challenge. To address this challenge, the real-world response was regarded as the reference response, while the simulation response was compensated to align with it. Both simulation and real-world experiments were conducted on a single shoulder joint, controlled using the manufacturer-provided PD controller. During the prompting phase, a $50^\circ$ target joint angle was specified, and a PD gain of $(250, 30)$ was applied in both simulation and real-world settings. In the test phase, an unseen target of $80^\circ$ was evaluated with altered PD gains of $(50, 8)$.

Crucially, unlike in Task~1 where the compensator directly augmented the control signal, the commercial platform does not permit modification of the control signal from PD controller. Therefore, the LLMs-guided compensator was used to operate indirectly by generating dynamic corrections to the target joint angle. By dynamically adjusting the target angle in simulation, the control signal from the PD controller was indirectly influenced, thereby modifying the resulting simulation response. 

\begin{figure*}
    \centering
    \includegraphics[width=1\linewidth]{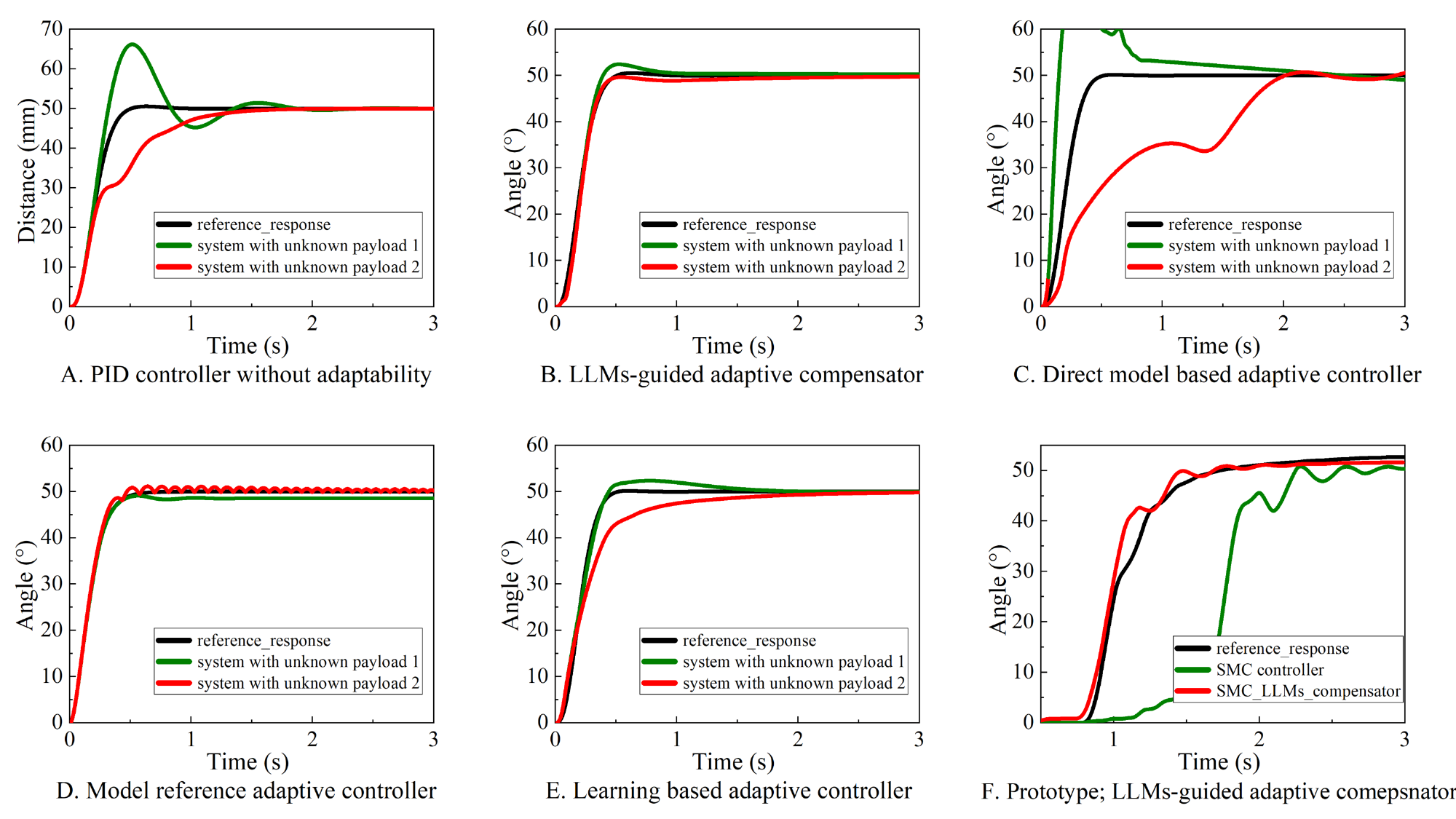}
    \caption{Response under different adaptive controller }
    \label{fig:enter-label}
\end{figure*}

\begin{figure}
    \centering
    \includegraphics[width=1\linewidth]{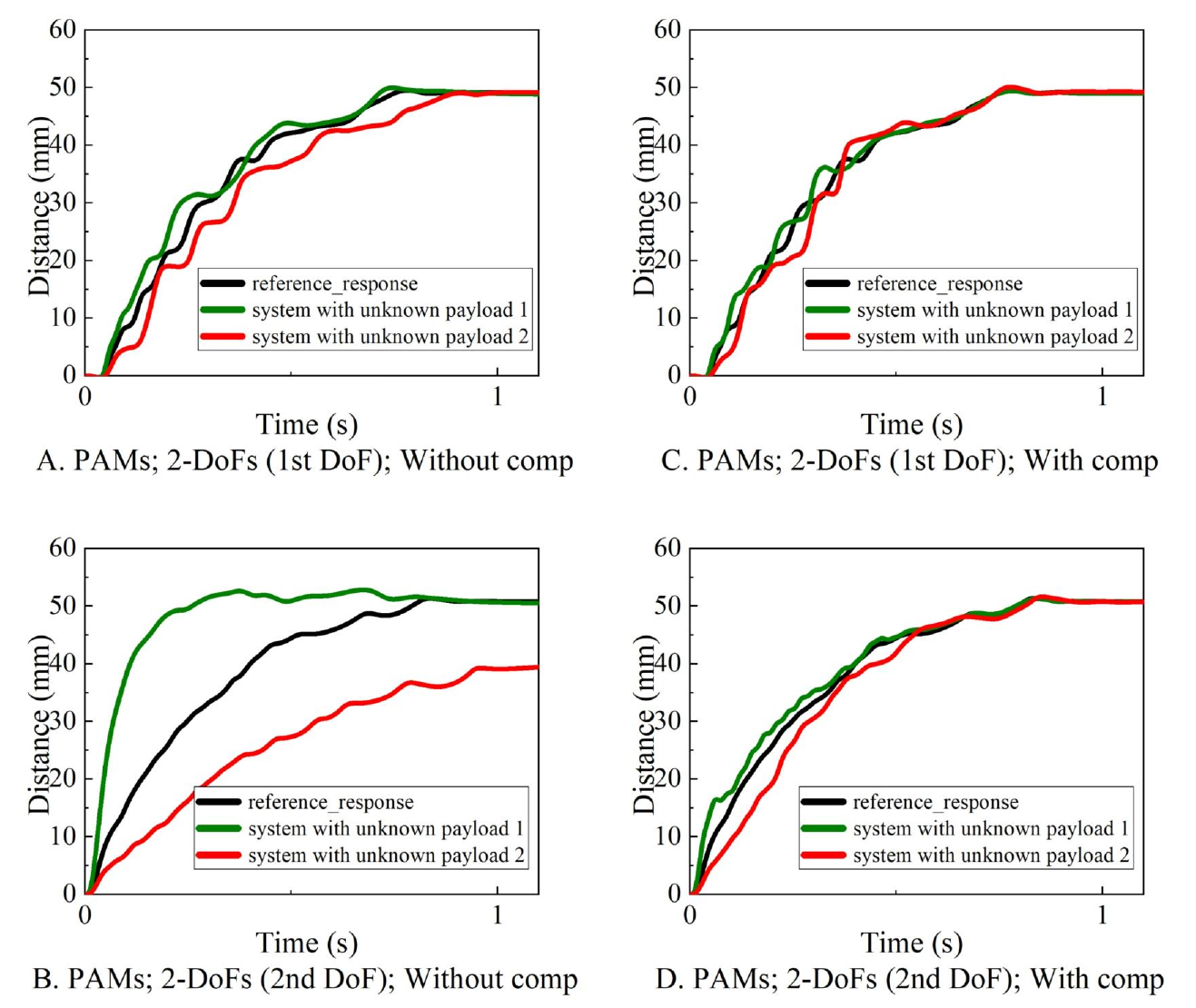}
    \caption{Response of 2 DoF PAMs-driven robot arm}
    \label{fig:enter-label}
\end{figure}

\begin{figure}
    \centering
    \includegraphics[width=1\linewidth]{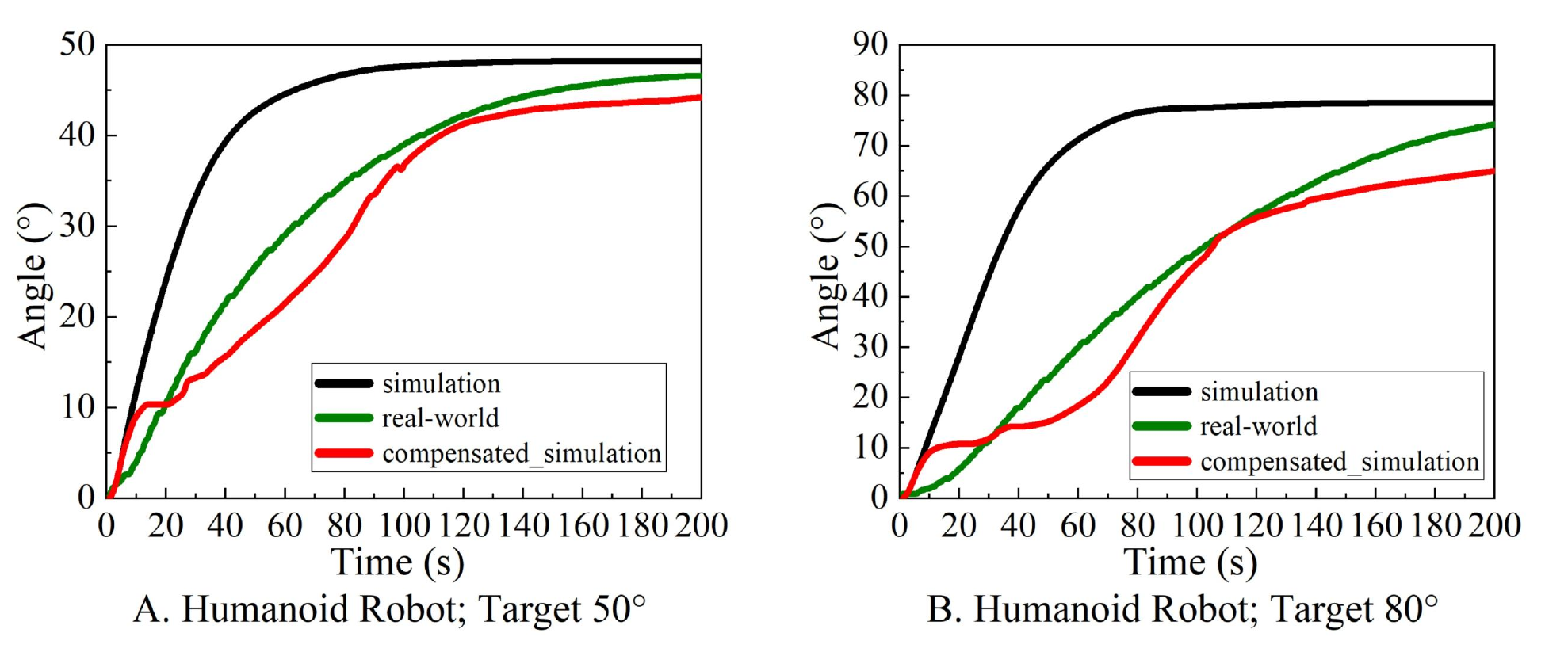}
    \caption{Response of Humanoid robot}
    \label{fig:enter-label}
\end{figure}

\section{RESULTS AND DISCUSSION}
\subsection{Results for Robot control test}
Figure 5 presents a comparative evaluation of the control performance of various controllers under different unknown payload conditions. Figure 5A shows the performance of a conventional PID controller. Due to the absence of any online adaptive capability, the system responses under the two different payloads vary significantly. The response exhibits large overshoot and discrepancies, indicating poor robustness of fixed-gain controllers when facing structural perturbations or parameter uncertainties.

Figure 5C illustrates the performance of a direct model-based adaptive controller. Although this controller theoretically guarantees asymptotic stability via Lyapunov methods, the Lyapunov function used only considers the state tracking error and payload estimation error. It does not incorporate transient performance metrics. As a result, under Payload 2, the system exhibits oscillatory behavior and delayed convergence. While it is theoretically possible to improve the performance by designing more sophisticated Lyapunov functions, this would significantly increase the design complexity and implementation burden.

Figure 5D presents the results of a MRAC. By explicitly incorporating a reference model into the control loop, the system is forced to follow a reference response. Consequently, it demonstrates good tracking performance under both payload conditions. However, the design of MRAC is often complex, as it typically requires structural assumptions, explicit reference models, and the derivation of adaptive laws. Moreover, both MRAC and direct model-based adaptive controllers fundamentally rely on accurate modeling and predefined structural assumptions. Their effectiveness depends on how precisely system dynamics and uncertainties are specified in advance. Although better model estimation generally yields better control performance in theory, it is often unclear in practice which components of the system are actually unknown. Moreover, many types of real-world uncertainties, including unmodeled dynamics, sensor biases, and actuator nonlinearities, are difficult to identify or incorporate into a fixed model structure. As a result, these model-driven approaches are inherently limited by assumptions that may not hold in realistic or unstructured environments.

Figure 5E shows the results of a learning-based adaptive controller (G-PID). This method does not require accurate robot's dynamic model. When sufficient data are available, it can achieve performance comparable to traditional controllers. Nevertheless, it suffers from several drawbacks, including heavy reliance on large-scale datasets, and instability during training. These issues are particularly pronounced when using adversarial learning structures, where training failure and unpredictable outcomes significantly hinder real-world deployment.

In contrast, Figure 5B demonstrates that the LLMs-guided adaptive compensator achieves the most stable and consistent performance among all tested methods. Under both payload conditions, the control strategy generated by the LLM is able to guide the system state smoothly and rapidly toward the reference trajectory, with virtually no overshoot or delay. Its key advantages include the complete elimination of explicit modeling, no need for deriving complex adaptive laws, and no dependence on large-scale data training. The entire compensator is generated from natural language task descriptions, reference trajectories, and observed system responses. The LLM is able to leverage its understanding of control principles and dynamic behavior to produce interpretable and executable compensator structures. The prototytpe experiment shown in Figure 5F further confirms the effectiveness of the proposed LLMs-guided adaptive compensator. This result represents one of multiple repeated trials, all of which consistently demonstrate that the compensator is capable of achieving reliable tracking performance under real-world conditions. 

It should be noted that, in theory, the performance of all adapitve controller could be further improved through extensive parameter tuning, structural modifications, or the addition of heuristic strategies. In the conducted experiments, reasonable tuning efforts were applied to each type of adapitve controller to achieve competitive results. However, optimal configurations were not guaranteed. Such tuning procedures are inherently labor-intensive and require significant expert intervention. By contrast, the LLMs-guided adaptive controller was designed and deployed without manual parameter adjustment, thereby demonstrating a high degree of usability.

Figure 6 validates the LLMs-guided adaptive compensator on a 2-DoF PAMs-driven robotic arm. Figures 6A and 6B show the uncompensated responses of the two joints under varying unknown payloads, revealing notable deviations, especially in the second DoF, including overshoot, delay, and failure to reach the target. These results highlight the challenges posed by payload variation and joint coupling.In contrast, Figures 6C and 6D show the compensated responses, where both DoFs closely follow the reference response with improved smoothness, faster convergence, and minimal steady-state error. Despite the second DoF being more challenging to control, its regulation does not noticeably degrade the first DoF’s performance, indicating that the compensator achieves effective task-aware decoupling, which is critical for coordinated control in multi-DoF PAMs systems.

Figure 7 presents the results of the LLMs-guided adaptive compensator on a humanoid robot platform. In both 50° and 90° cases, the compensator successfully adjusted the simulation response to closely match the real -world system response.  It is worth noting that the compensated simulation exhibits oscillations at the early stage of motion. Due to the closed architecture of the humanoid controller, the compensator can only adjust the desired angle indirectly rather than directly modifying the control signal. This introduces an additional nonlinear mapping in the control pathway, significantly increasing the difficulty of effective compensation.Moreover, because the first-step states of the simulation and the real system are nearly identical, the LLM model implicitly treats this initial phase as requiring no compensation.  Despite these structural and informational limitations, the compensator still demonstrates effective alignment between simulation and real-world behavior, highlighting its adaptability and practical potential under constrained control interfaces.

\subsection{Analyze of Reasoning path of LLMs}
During the Type 1, 1-DoF PAM-driven robot arm experiments, the reasoning paths followed by LLMs were analyzed, as summarized in Table~\ref{tab:llm_reasoning}. Although both frameworks aim to achieve adaptive control, distinctly different reasoning approaches were exhibited by the LLMs depending on the design task.\\

\begin{table}[ht]
\caption{Reasoning path of two LLMs-based automatic controller}
\label{tab:llm_reasoning}
\setlength{\tabcolsep}{4pt}
\renewcommand{\arraystretch}{1.5}
\begin{tabular}{@{}p{20pt}p{220pt}@{}}
\specialrule{2pt}{0pt}{0pt}  
\textbf{Step} & \textbf{Reasoning trace for Compensator design} \\
\midrule
1 & Extract step-response metrics for all systems \\
2 & Error analysis: System 1 $\rightarrow$ overshoot; System 2 $\rightarrow$ slow \\
3 & Characterize the mismatch: System 1 implies under-damping; a derivative boost or lead compensation might help improve transient behavior. System 2 suggests a static gain mismatch, which requires a compensation term with integral action or low-frequency gain correction. \\
4 & Compensator proposal \\
\midrule
\multicolumn{2}{@{}p{240pt}@{}}{\textbf{Indicators:} Peak overshoot, Settling time, Steady-state error, Max/Min error, etc.} \\
\specialrule{2pt}{0pt}{0pt}  
\end{tabular}
\end{table}

\begin{table}[ht]
\caption{Reasoning path of adaptive-enhanced controller}
\label{tab:adaptive_reasoning}
\setlength{\tabcolsep}{4pt}
\renewcommand{\arraystretch}{1.5}
\begin{tabular}{@{}p{20pt}p{220pt}@{}}
\specialrule{2pt}{0pt}{0pt}  
\textbf{Step} & \textbf{Reasoning trace for Adaptive-enhanced controller} \\
\midrule
1 & Error definition: $e(t) = r(t) - y(t)$ \\
2 & Cost function selection: $J(t) = \frac{1}{2}e^2$. A gradient or Lyapunov approach then tells us how to nudge the gains. \\
3 & Choose an adaptation law: The classic MIT rule: $\dot{\theta} = -\gamma \frac{\partial J}{\partial \theta}$. Because $J = \frac{1}{2} e^2$, we need $\partial e / \partial \theta$. Assuming the closed loop is minimum phase and $\partial e / \partial u \approx -1/\lambda$ (a positive constant), we get simple update rules. \\
4 & Practical considerations: Anti-windup, derivative filtering, discrete-time implementation, boundedness. \\
5 & Compensator proposal \\
\midrule
\multicolumn{2}{@{}p{240pt}@{}}{\textbf{Indicators:} Stability guarantee, adaptive-law performance, transient quality, parameter boundedness, closed-loop poles, etc.} \\
\specialrule{2pt}{0pt}{0pt}  
\end{tabular}
\end{table}

\begin{table}[ht]
\centering
\caption{Performance comparison on three selected tasks (success rate / iteration count)}
\label{tab:combined_table}
\setlength{\tabcolsep}{10pt}
\renewcommand{\arraystretch}{1.5}
\begin{tabular}{@{}lccc@{}}
\specialrule{1.2pt}{0pt}{0pt}
 & \textbf{T-1 1-DoF} & \textbf{T-1 2-DoF} & \textbf{T-2} \\
\midrule
Adaptive Controller   & 20\% / 10.2 & 0\% / ---   & 10\% / 6.2 \\
Adaptive Compensator  & 80\% / 5.4  & 70\% / 8.6  & 60\% / 8.4 \\
\specialrule{1.2pt}{0pt}{0pt}
\end{tabular}
\end{table}

In the LLMs-guided adaptive compensator, as illustrated in Table~\ref{tab:llm_reasoning}, LLMs primarily rely on system-level feature analysis to design compensator. By comparing the step responses of the unknown and reference systems, they identify discrepancies in dynamic behavior and infer underlying physical characteristics. Based on these inferences, they propose appropriate compensation strategies to improve transient or steady-state performance.

This reasoning process reflects a clear CoT pattern:

\noindent
\textbf{Observe step response} $\rightarrow$ \textbf{Identify performance issues} (e.g., overshoot or slow response) $\rightarrow$ \textbf{Infer physical mechanisms} (e.g., under-damping or low gain) $\rightarrow$ \textbf{Derive compensation direction} (e.g., derivative, integral, or feedforward) $\rightarrow$ \textbf{Design and tune the compensator}

Although system responses can result from a variety of underlying factors, LLMs are still able to reason through a clear logical chain. Even when initial assumptions are not entirely accurate, the models can progressively converge toward the root cause through iterative interactions and refinements. Therefore, this reasoning strategy is primarily driven by observable system responses and focuses on key performance metrics such as peak overshoot, settling time, and steady-state error, thereby guiding control design in a structured and interpretable manner.

In contrast, within the LLMs-guided adaptive controller, LLMs tend to adopt a model-based reasoning approach rooted in control theory. The process typically starts with defining the error function, constructing a performance index, and deriving adaptation laws via Lyapunov or gradient-based methods.
This reasoning process is distinctly theory-driven, and its CoT can be summarized as follows:

\noindent
\textbf{Define error function} $\rightarrow$ \textbf{Construct Lyapunov function} $\rightarrow$ \textbf{Introduce system assumptions} (e.g., minimum phase) $\rightarrow$ \textbf{Derive adaptation law} (e.g., MIT rule) $\rightarrow$ \textbf{Account for practical constraints} (e.g., integrator saturation, derivative filtering, parameter bounds) $\rightarrow$ \textbf{Implement adaptation law to update controller parameters}

LLMs also consider key implementation aspects such as anti-windup, filtering, discretization, and parameter bounds. This reasoning pathway aligns closely with traditional control-theoretic design, emphasizing internal controller structure, stability verification, and robustness—reflecting a theory-driven and systematic approach.

\section{Conclusion}
In this work, we investigated whether LLMs can  contribute to the domain of automatic control. We proposed an LLMs-guided adaptive compensator that enhances existing feedback controllers by steering their control response toward that of a predefined reference response, thereby achieving adaptivity without requiring explicit system modeling. To validate the proposed compensator, we conducted systematic comparisons against traditional adaptive controller across both simulation and prototype experiments, including multiple robotic systems with varying complexity. The results demonstrate that the LLMs-guided compensator not only achieves superior control performance but also significantly reduces design effort and system-specific tuning requirements. Furthermore, through Lyapunov-based stability analysis and reasoning-path investigations, we demonstrate that this zero-shot prompting strategy exhibits strong generalization, adaptivity, and robustness

As LLMs continue to gain traction in robotics, compared to VLM-based methods, which often require retraining from scratch or replacing the entire control system, prompt-driven compensator design offers a lightweight, safety, and non-intrusive alternative. This makes it particularly advantageous for legacy or already-deployed robotic platforms where modifying existing controllers is infeasible or costly.



\section*{Acknowledgment}

This work acknowledges the contributions of Mr. Tianyi Yang and Dr. Qian Niu for their assistance with illustrations and conceptual discussions.




\end{document}